\documentclass[runningheads]{llncs}
\usepackage[T1]{fontenc}
\usepackage{graphicx,verbatim}
\usepackage{cite}
\usepackage{amsmath, amssymb}
\usepackage{multirow}
\usepackage{pifont}
\usepackage{booktabs}
\usepackage{marvosym}
\begin{document}
\title{CA-GCL: Cross-Anatomy Global-Local Contrastive Learning for Robust 3D Medical Image Understanding}
\titlerunning{CA-GCL: Cross-Anatomy Global-Local Contrastive Learning}
% \begin{comment}  %% Removed for anonymized MICCAI submission

% \end{comment}

\author{
Hanwen Zhang\inst{1} \and
Yao Liu\inst{1} \and
Die Dai\inst{1} \and
Jiaye Yang\inst{1} \and
Qiao Liu\inst{1} \and
Yutong Xie\inst{2} \and
Peng Wang\inst{1}
}

\authorrunning{H. Zhang et al.}

\institute{
University of Electronic Science and Technology of China, Chengdu, China \and
Mohamed bin Zayed University of Artificial Intelligence, Abu Dhabi, UAE \\
\Letter \ \email{202522080828@std.uestc.edu.cn}
}

\maketitle              % typeset the header of the contribution
\begin{abstract}
Fine-grained Vision-Language Pre-training (FVLP) demonstrates significant potential in 3D medical image understanding by aligning anatomy-level visual representations with corresponding textual descriptions. However, existing FVLP paradigms often suffer from severe representation collapse in the textual embedding space, where text embeddings of distinct anatomical structures become highly clustered and indistinguishable. This distributional degeneracy renders the model hypersensitive to prompt variations, hindering reliable clinical deployment. To address these challenges, we propose a novel \textbf{C}ross-\textbf{A}natomy \textbf{G}lobal-Local \textbf{C}ontrastive \textbf{L}earning framework (\textbf{CA-GCL}). CA-GCL introduces a global contrastive objective that enforces separation between anatomical categories in the latent space, effectively counteracting the aggregation tendency induced by local alignment. Furthermore, we incorporate a clinical-aware text augmentation strategy based on permutation invariance and partial completeness to enhance robustness against descriptive incompleteness. Extensive evaluations on the CT-RATE and Rad-ChestCT datasets show that CA-GCL achieves comparable zero-shot abnormality detection performance to existing VLP paradigms, while demonstrating substantially better robustness to prompt variations: on canonical templates it obtains higher mean AUC with lower variance, and on non-canonical templates it remains stable whereas baselines degrade markedly. These results validate CA-GCL as an effective framework for robust 3D medical image understanding. Code is available at \url{https://github.com/W11H08Z/CA-GCL}.

% 细粒度视觉-语言预训练（FVLP）通过将解剖级别的视觉表征与相应的文本描述进行对齐，在三维医学图像理解方面展现出巨大潜力。然而，现有的FVLP范式在文本嵌入空间中往往存在严重的表征崩溃问题，即不同解剖结构的文本嵌入变得高度聚集且难以区分。这种分布退化导致模型对提示词变化极为敏感，阻碍了其在临床环境中的可靠部署。为解决这些问题，我们提出了一种新颖的跨解剖全局-局部对比学习框架（CA-GCL）。CA-GCL引入了一种全局对比目标，强制潜在空间中不同解剖类别之间的分离，有效抵消了局部对齐所导致的聚合趋势。此外，我们基于排列不变性和部分完整性设计了一种临床感知的文本增强策略，以增强模型对描述不完整的鲁棒性。在CT-RATE和Rad-ChestCT数据集上的大量评估表明，CA-GCL在零样本异常检测中始终优于现有的视觉-语言预训练范式，在实现卓越性能的同时展现出强大的跨数据集泛化能力。关键的是，CA-GCL降低了不同提示词模板下的性能方差，将原本崩溃的文本相似度分布转化为结构良好的流形。这些结果验证了CA-GCL作为鲁棒三维医学图像理解的有效框架。

% Authors must provide keywords and are not allowed to remove this Keyword section.

\end{abstract}
\section{Introduction}

3D medical image understanding via volumetric modalities like CT and MRI is vital for clinical diagnosis, providing anatomical context and spatial continuity. While early supervised methods~\cite{isensee2021nnu, wang2025sam, zhang2023parse, xie2021cotr, zhang2020inter} achieved strong performance, they suffer from limited scalability due to high expert annotation costs. Vision-language pre-training (VLP)~\cite{zhang2022contrastive, lin2023pmc, hamamci2024foundation, blankemeier2024merlin, huang2021gloria, wang2022multi} has emerged as a scalable alternative, leveraging paired images and radiology reports to learn transferable representations for zero-shot 3D medical image understanding~\cite{ni2024mg, bai2024m3d, wu2025vision}.

\begin{figure}[t]
\centering
\includegraphics[width=\textwidth]{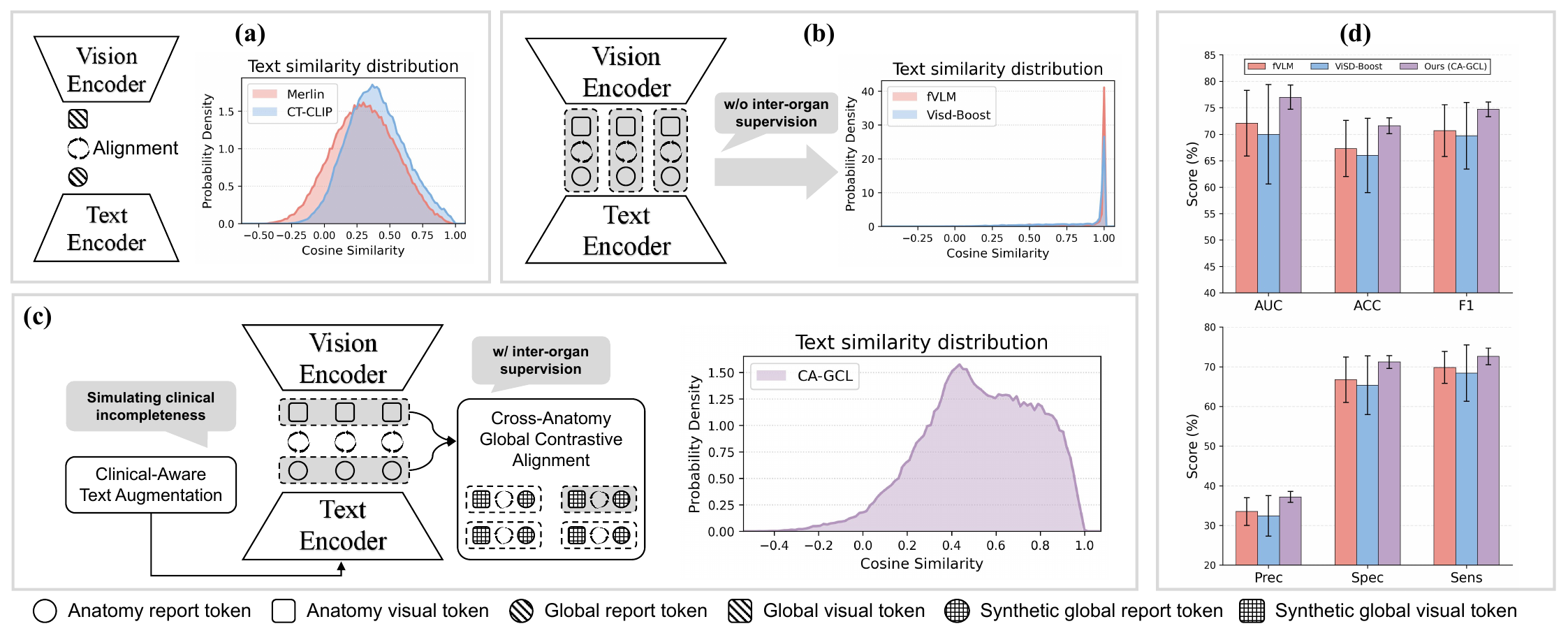}
\caption{Comparison of VLP paradigms and their corresponding text similarity distributions. (a) Global VLP aligns global visual and report tokens. (b) Fine-grained VLP performs pairwise anatomical alignment but suffers from severe distributional degeneracy (similarity peak near $1.0$). (c) CA-GCL (Ours) aggregates anatomy-specific representations into synthetic global tokens while maintaining local alignment, effectively mitigating representation collapse to ensure a discriminative latent space. (d) Zero-shot robustness comparison on CT-RATE across 10 prompt templates (Table~\ref{tab:prompt_groups}).}
\label{fig:1}
\end{figure}

To further improve alignment quality, recent studies have advanced from Global VLP (Fig.~\ref{fig:1}(a)) to FVLP, where images and reports are decomposed into anatomy-level units (Fig.~\ref{fig:1}(b)). By explicitly aligning visual representations of individual organs with their corresponding report, FVLP methods exhibit superior performance in zero-shot classification tasks~\cite{lin2024ct, shui2025large, cao2025boosting}. However, existing FVLP methods suffer from representation collapse in the textual embedding space. First, since visual appearances of the same organ are highly similar across patients, where pathologies often occupy only a minimal volume, visual representations naturally cluster. Second, fine-grained alignment inevitably transfers this intra-organ clustering to the textual modality, leading to highly concentrated embeddings for identical anatomical categories. Moreover, current FVLP frameworks lack explicit supervision to distinguish between textual representations of different anatomical structures. Consequently, the text encoder tends to learn a "shortcut" where textual embeddings associated with distinct organs are progressively drawn closer, inducing distributional degeneracy (Fig.~\ref{fig:1}(b)). This renders the model's zero-shot abnormality detection performance hypersensitive to prompt templates, causing substantial performance fluctuations when prompts deviate from the training distribution, which poses a significant barrier to reliable clinical deployment.

To address these limitations, we propose a novel training paradigm designed to explicitly regularize the geometry of fine-grained vision–language representations (Fig.\ref{fig:1}(c)), namely \textbf{C}ross-\textbf{A}natomy \textbf{G}lobal-Local \textbf{C}ontrastive \textbf{L}earning (\textbf{CA-GCL}), which consists of two core modules: Cross-Anatomy Global Contrastive Alignment and Clinical-Aware Text Augmentation. The Cross-Anatomy Global Contrastive Alignment module computes a global visual token by averaging randomly sampled visual tokens across all anatomical regions and a corresponding global report token by averaging the associated anatomy-level report tokens. Since global visual tokens aggregate information from multiple distinct organs with low similarity, their distribution is naturally dispersed. In contrast, prior to optimization, the global report tokens remain concentrated due to the high similarity between individual organ reports. By aligning these global pairs, the global report tokens are forced to diversify. To maintain local alignment quality, the model is compelled to "shift" the entire clusters of different organs away from each other, effectively forcing the separation of anatomy-specific report tokens (Fig.~\ref{fig:4}). Furthermore, we incorporate a Clinical-Aware Text Augmentation strategy that utilizes random shuffling and subset sampling to simulate descriptive variations and report incompleteness, thereby enhancing the model's robustness against descriptive incompleteness and noise.

In summary, our contributions are threefold: (1) We propose CA-GCL, a global-local contrastive paradigm that integrates cross-anatomy objectives to mitigate representation collapse and ensure embedding discriminability. (2) We introduce a clinical-aware text augmentation strategy based on permutation invariance and partial completeness, enhancing robustness against descriptive incompleteness and noise. (3) We demonstrate that CA-GCL achieves comparable zero-shot performance to existing methods on CT-RATE and Rad-ChestCT datasets, while exhibiting substantially better robustness to prompt variations with significantly lower variance.

\section{Method}
\label{sec:method}

\subsection{Anatomy-aware Local Contrastive Alignment}

We denote a dataset with $N$ paired image-report samples as
$\mathcal{D} = \{(X_i^I, X_i^R)\}_{i=1}^N$,
where $X_i^I$ denotes the $i$-th CT image and $X_i^R$ is its corresponding radiology report.
For each CT image, we apply a whole-body segmentation model~\cite{wasserthal2023totalsegmentator} to decompose it into
$M$ anatomical regions,
$X_i^I \rightarrow \{X_{i,1}^I, \ldots, X_{i,M}^I\}.$
On the textual side, we employ \texttt{deepseek-chat}~\cite{guo2025deepseek} to perform structured parsing of the entire report. Unlike~\cite{shui2025large}, which parses the \emph{Findings} and \emph{Impression} sections separately, we adopt Report Jointly Parsing for both sections to mitigate semantic inconsistencies and contradictions inherent in segmented processing.
After parsing, each report is represented as a set of anatomy-level report
corresponding to individual organs,
$X_i^R \rightarrow \{X_{i,1}^R, \ldots, X_{i,M}^R\}.$

Following~\cite{shui2025large}, we adopt ViT~\cite{dosovitskiy2020image} and BERT~\cite{boecking2022making, devlin2019bert} as the visual and textual encoders,
respectively.
Given the sequence of visual tokens produced by ViT, anatomy-level visual tokens
$f_{i,j}^I$ are extracted according to the corresponding organ masks.
On the text side, the anatomy-level report $X_{i,j}^R$ is directly fed into
the text encoder to obtain anatomy-level report tokens $f_{i,j}^R$.
To aggregate the anatomy-level visual tokens, we introduce a learnable query token
$Q_{i,j}^I$ for each organ and perform cross-attention,  yielding the anatomy visual token:
$
V_{i,j} = \mathrm{CrossAttn}(Q_{i,j}^I, f_{i,j}^I, f_{i,j}^I).
$
For the textual modality, the \texttt{[CLS]} token of $f_{i,j}^R$ is used as the anatomy report token $R_{i, j}$.

For the $j$-th anatomical structure, let $N_j$ denote the number of samples with complete
anatomical structures within a mini-batch.
The anatomy-aware local contrastive loss~\cite{oord2018representation} is formulated in a bidirectional manner as
\begin{equation}
\begin{aligned}
\mathcal{L}_{\mathrm{loc}}^{(j)}
=
- \frac{1}{N_j} \sum_{i=1}^{N_j}
\left(\log
\frac{e^{\langle V_{i,j}, R_{i,j} \rangle / \tau }}
{\sum_{k=1}^{N_j} e^{\langle V_{i,j}, R_{k,j} \rangle / \tau }}
 + \log
\frac{e^{\langle V_{i,j}, R_{i,j} \rangle / \tau }}
{\sum_{k=1}^{N_j} e^{\langle V_{k,j}, R_{i,j} \rangle / \tau }}
\right),
\end{aligned}
\end{equation}
where $\langle \cdot, \cdot \rangle$ denotes cosine similarity and $\tau$ is a learnable temperature coefficient, optimized during training to adaptively scale the similarity scores.

% The overall anatomy-aware local contrastive loss is obtained by summing over all
% anatomical structures:
% \begin{equation}
% \mathcal{L}_{\mathrm{loc}}
% =
% \sum_{j=1}^{M}
% \mathcal{L}_{\mathrm{loc}}^{(j)}.    
% \end{equation}

\begin{figure}[t]
\centering
\includegraphics[width=\textwidth]{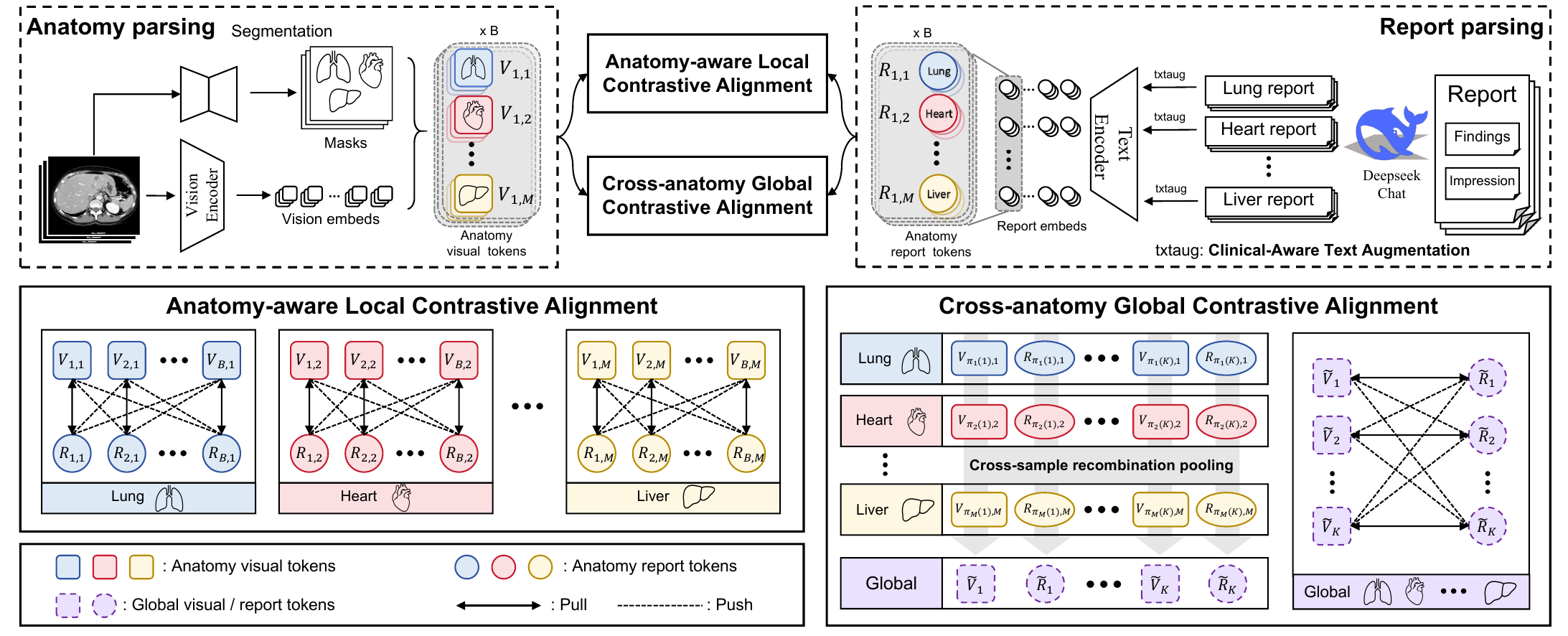}
\caption{The pipeline of our proposed CA-GCL framework. The framework extracts anatomy-level tokens from CT images and radiology reports (Top). It integrates Anatomy-aware Local Contrastive Alignment for organ-specific matching (Bottom-Left) and Cross-anatomy Global Contrastive Alignment (Bottom-Right).}
\label{fig:2}
\end{figure}

\subsection{Cross-Anatomy Global Contrastive Alignment}

% In 3D medical imaging, visual representations of the same anatomical structure naturally form clusters in the latent space due to their inherent morphological similarities. While local alignment effectively bridges the modalities, it inadvertently propagates this clustering effect to the textual manifold, pulling textual embeddings into highly concentrated regions. Without explicit supervisory signals to enforce mutual repulsion between distinct organs, the textual representations suffer from feature collapse, undermining their discriminability. To counteract this undesired aggregation, we propose a cross-anatomy global contrastive alignment module to regularize the global geometry of the joint embedding space.

In a mini-batch of size $B$, we obtain $M$ anatomy-level image-text pairs for each sample:
$
\{(V_{i,1}, R_{i,1}), \ldots, (V_{i,M}, R_{i,M})\}_{i=1}^B.
$
Directly aligning these pairs within a single CT scan may introduce representation bias due to the fixed co-occurrence of organs in a patient. To break this dependency and enrich the diversity of global representations, we perform cross-sample recombination at the batch level. 

For each anatomical structure $j \in \{1,\ldots,M\}$, we have a set of representations from different patients: $\{(V_{1,j}, R_{1,j}), \ldots, (V_{B,j}, R_{B,j})\}$.
We then generate $K$ new "synthetic" global sample groups by randomly shuffling and recombining these anatomical components across different patients. The $k$-th recombined global sample group $\mathcal{G}_k$ is defined as:
\begin{equation}
\mathcal{G}_k = \big( (V_{\pi_1(k),1}, R_{\pi_1(k),1}), \ldots, (V_{\pi_M(k),M}, R_{\pi_M(k),M}) \big), \quad k=1,\ldots,K,
\end{equation}
where $\pi_j$ is a random permutation of the patient indices $\{1,\ldots,B\}$. This shuffling strategy ensures that each anatomy-level representation is utilized exactly once across all global sample groups, preventing data redundancy. After recombination, we derive the synthetic global image and text representations:
\begin{equation}
\tilde{V}_k = \frac{1}{M}\sum_{j=1}^M V_{\pi_j(k),j}, \quad \tilde{R}_k = \frac{1}{M}\sum_{j=1}^M R_{\pi_j(k),j}.
\end{equation}
where $k=1,\ldots,K$. Finally, we impose a cross-anatomy global contrastive loss on these recombined pairs $\{(\tilde{V}_k, \tilde{R}_k)\}_{k=1}^K$:
\begin{equation}
\mathcal{L}_{\text{glo}} = - \frac{1}{K} \sum_{i=1}^{K} \left( \log \frac{e^{\langle \tilde{V}_i,\tilde{R}_i \rangle / \tau}}{\sum_{j=1}^{K} e^{\langle \tilde{V}_i,\tilde{R}_j \rangle / \tau}} + \log \frac{e^{\langle \tilde{V}_i,\tilde{R}_i \rangle / \tau}}{\sum_{j=1}^{K} e^{\langle \tilde{V}_j,\tilde{R}_i \rangle / \tau}} \right).
\end{equation}

The total training objective is defined as a weighted combination of the local and global alignment losses:
\begin{equation}
\mathcal{L}_{\text{total}} = \sum_{j=1}^{M}\mathcal{L}_{\mathrm{loc}}^{(j)} + \lambda \mathcal{L}_{\text{glo}},
\end{equation}
where $\lambda = 0.1$ is a balancing hyperparameter. 

\subsection{Clinical-Aware Text Augmentation}

We introduce a text augmentation strategy for each anatomy-level report $X^R_{i,j}$. Our approach is grounded in two clinical observations: (1) \emph{Permutation Invariance}: The semantic integrity of a report is largely preserved regardless of the descriptive order of anatomical findings. (2) \emph{Partial Completeness}: Clinicians often omit mentions of normal organs or simplify non-diagnostic regions. Specifically, we decompose each report $X^R_{i,j}$ into a set of sentences $\{s_1, \ldots, s_{L_{i,j}}\}$. We then generate an augmented version $\tilde{X}^R_{i,j} = \mathrm{Concat}(\tilde{\mathcal{S}}_{i,j})$ by randomly shuffling and sampling a non-empty subset $\tilde{\mathcal{S}}_{i,j} \subseteq \{s_1, \ldots, s_{L_{i,j}}\}$. This perturbed representation $\tilde{X}^R_{i,j}$ is then aligned with the corresponding visual feature $V_{i,j}$ to regularize the model against descriptive incompleteness and noise. 

\section{Experiments and Results}

\subsection{Datasets and Evaluation Metrics}

We evaluate our method using two large-scale CT datasets: CT-RATE~\cite{hamamci2024foundation} and Rad-ChestCT~\cite{draelos2021machine}. The model is trained on the official training split of CT-RATE. For performance evaluation, we conduct internal validation on the CT-RATE test set and assess external generalization across the Rad-ChestCT dataset, which comprises 3,630 CT volumes. We compare the performance of different VLP paradigms through zero-shot abnormality detection. Following~\cite{shui2025large}, performance is quantified using the Area Under the Receiver Operating Characteristic curve (AUC), balanced accuracy (ACC), specificity (Spec), sensitivity (Sens), precision (Prec), and the weighted F1-score. Furthermore, we evaluate the geometric properties of the representation space by t-SNE~\cite{van2008visualizing} visualizations of the joint image-text embedding distributions.

\subsection{Implementation Details}

\subsubsection{Image Pre-processing.} 
We employ TotalSegmentator~\cite{wasserthal2023totalsegmentator} to extract 104 organ structures from each CT scan and aggregate them into $M=36$ primary anatomies following~\cite{shui2025large}. All volumes are resampled to $1\times1$~mm in-plane resolution with $3$~mm slice thickness. Hounsfield Unit values are clipped to $[-1150, 350]$ and linearly normalized to $[0, 1]$. During training, we randomly crop sub-volumes of size $96\times256\times384$ voxels (axial, coronal, sagittal) and retain only anatomies fully contained within each crop. To prevent partial anatomical cues from misleading the contrastive learning process, we restrict supervision to organs that are completely contained within the current crop, disregarding any structures truncated by the cropping boundary.

\subsubsection{Encoder Initialization.} Following~\cite{shui2025large}, we adopt a vision transformer (ViT-base)~\cite{dosovitskiy2020image} as the image encoder, initialized with the same MAE ImageNet-1K pre-trained weights~\cite{he2022masked}. The patch size is set to $16 \times 16 \times 32$ along the axial, coronal, and sagittal axes, respectively. A pre-trained BERT model~\cite{boecking2022making} is employed as the text encoder, consistent with the checkpoint used in~\cite{shui2025large}.

\subsubsection{Training Configuration.} The model is trained for 20 epochs on 4 NVIDIA H100 GPUs with a total batch size of 56. We use the Adam optimizer with a learning rate schedule consisting of two phases: a linear warmup from $0$ to $1\mathrm{e}^{-4}$ over the first epoch, followed by cosine annealing that decays the learning rate from $1\mathrm{e}^{-4}$ to $1\mathrm{e}^{-6}$ over the remaining 19 epochs. RandomCrop and RandomFlip are applied on the fly. We use uniform sampling to ensure at least one anatomy is fully included in each crop.

\subsection{Overall Performance}

\subsubsection{Zero-shot Performance.}
We report the zero-shot abnormality detection performance on the CT-RATE dataset. To ensure a fair and consistent comparison, all baseline results are directly obtained from the original publications. As shown in Table~\ref{tab:exp1_best}, CA-GCL achieves highly competitive results among all zero-shot VLP baselines. Specifically, our method attains an AUC of 79.3, Specificity of 71.9, and Sensitivity of 75.5, while remaining comparable to strong baselines on ACC, F1, and Precision. Notably, CA-GCL surpasses several supervised fine-tuned methods (e.g., CT-VocabFine and CT-LiPro) in terms of AUC and F1, demonstrating the effectiveness of the proposed framework.

\begin{table}[t]
  \centering
  \scriptsize
  \setlength{\tabcolsep}{2.5pt}
  \caption{Zero-shot performance on the CT-RATE and Rad-ChestCT datasets. The best result in each column is \textbf{bold}, and the second best is \underline{underlined}.}
  \begin{tabular}{lcccccccccccc}
    \toprule
    \multirow{2}{*}{Method} & \multicolumn{6}{c}{CT-RATE} & \multicolumn{6}{c}{Rad-ChestCT} \\
    \cmidrule(lr){2-7} \cmidrule(lr){8-13}
    & AUC & ACC & F1 & Prec & Spec & Sens & AUC & ACC & F1 & Prec & Spec & Sens \\
    \midrule
    \emph{Supervised} & & & & & & & & & & & & \\
    CT-VocabFine & 76.0 & 70.4 & 73.8 & \underline{35.6} & -- & -- & 65.7 & 62.1 & 66.8 & \underline{35.6} & -- & -- \\
    CT-LiPro & 76.1 & 69.1 & 72.6 & 34.3 & -- & -- & 64.7 & 60.6 & 65.0 & 35.1 & -- & -- \\
    \midrule
    \emph{Zero-shot} & & & & & & & & & & & & \\
    CT-CLIP~\cite{hamamci2024foundation} & 73.3 & 66.9 & 70.8 & 32.6 & -- & -- & 63.2 & 59.9 & 64.7 & 34.1 & -- & -- \\
    BIUD~\cite{cao2024bootstrapping} & 71.3 & 68.1 & 71.6 & 33.8 & 68.6 & 67.3 & 62.9 & 60.6 & 65.2 & 33.7 & 60.2 & 59.6 \\
    Merlin~\cite{blankemeier2024merlin} & 72.8 & 67.2 & 70.9 & 33.7 & 66.8 & 70.1 & 64.4 & 61.9 & 66.3 & 34.8 & 61.7 & 61.0 \\
    fVLM~\cite{shui2025large} & 77.8 & 71.8 & 75.1 & 37.9 & \underline{71.7} & \underline{72.8} & 68.0 & 64.7 & 68.8 & \textbf{37.4} & \underline{64.6} & \underline{64.6} \\
    ViSD-Boost~\cite{cao2025boosting} & \underline{79.0} & \textbf{73.1} & \textbf{75.9} & \textbf{38.7} & -- & -- & \textbf{69.4} & \textbf{65.2} & \textbf{69.3} & 34.2 & -- & -- \\
    Ours & \textbf{79.3} & \underline{72.7} & \underline{75.7} & \underline{38.4} & \textbf{71.9} & \textbf{75.5} & \underline{69.2} & \underline{64.7} & \underline{69.0} & \underline{35.1} & \textbf{65.2} & \textbf{64.8} \\
    \bottomrule
  \end{tabular}
\label{tab:exp1_best}
\end{table}

\subsubsection{Robustness to Prompt Variations.}
To assess clinical deployability, we evaluate zero-shot stability across prompt templates (Table~\ref{tab:prompt_groups}). We partition the 10 templates into two groups based on their linguistic conformity to standard radiological reporting conventions; the grouping criteria and template lists are detailed in Appendix~\ref{app:prompt_grouping}. Mean and standard deviation within each group are reported in Table~\ref{tab:prompt_canonical} and Table~\ref{tab:prompt_noncanonical}, while the complete per-template breakdown is provided in Appendix~\ref{app:per_template}.

\begin{table*}[t]
  \centering
  \setlength{\tabcolsep}{2pt}
  \caption{Zero-shot performance on \textbf{Canonical} prompt templates. Results are reported as mean$\pm$std across 5 templates. The best result in each column is \textbf{bold}.}
  \begin{tabular}{lcccccc}
    \toprule
    Method & AUC & ACC & F1 & Prec & Spec & Sens \\
    \midrule
    \multicolumn{7}{c}{\textbf{CT-RATE}} \\
    \midrule
    CT-CLIP~\cite{hamamci2024foundation} & 71.9$\pm$1.0 & 66.1$\pm$0.8 & 70.1$\pm$0.7 & 31.9$\pm$0.6 & 65.3$\pm$0.9 & 69.9$\pm$0.6 \\
    fVLM~\cite{shui2025large} & 76.4$\pm$1.8 & 70.5$\pm$1.5 & 73.6$\pm$1.4 & 36.0$\pm$1.1 & 70.1$\pm$1.4 & 72.8$\pm$1.2 \\
    ViSD-Boost~\cite{cao2025boosting} & 76.1$\pm$2.7 & 70.2$\pm$2.4 & 73.5$\pm$2.0 & 35.5$\pm$2.5 & 69.5$\pm$2.3 & 72.6$\pm$2.7 \\
    Ours & \textbf{78.2$\pm$0.7} & \textbf{72.5$\pm$0.5} & \textbf{75.5$\pm$0.4} & \textbf{38.0$\pm$0.5} & \textbf{72.2$\pm$0.7} & \textbf{73.5$\pm$1.1} \\
    \midrule
    \multicolumn{7}{c}{\textbf{Rad-ChestCT}} \\
    \midrule
    fVLM~\cite{shui2025large} & 67.0$\pm$1.4 & 63.2$\pm$1.5 & 67.7$\pm$1.3 & 33.0$\pm$1.1 & 62.9$\pm$1.8 & 64.2$\pm$1.3 \\
    ViSD-Boost~\cite{cao2025boosting} & 66.7$\pm$3.2 & 62.2$\pm$2.7 & 66.8$\pm$2.4 & 32.0$\pm$2.4 & 62.1$\pm$2.6 & \textbf{64.2$\pm$3.0} \\
    Ours & \textbf{67.2$\pm$1.2} & \textbf{64.0$\pm$0.9} & \textbf{68.3$\pm$0.8} & \textbf{33.9$\pm$0.6} & \textbf{63.8$\pm$1.1} & 63.7$\pm$1.7 \\
    \bottomrule
  \end{tabular}
\label{tab:prompt_canonical}
\end{table*}

\begin{table*}[t]
  \centering
  \setlength{\tabcolsep}{2pt}
  \caption{Zero-shot performance on \textbf{Non-canonical} prompt templates. Results are reported as mean$\pm$std across 5 templates. The best result in each column is \textbf{bold}.}
  \begin{tabular}{lcccccc}
    \toprule
    Method & AUC & ACC & F1 & Prec & Spec & Sens \\
    \midrule
    \multicolumn{7}{c}{\textbf{CT-RATE}} \\
    \midrule
    CT-CLIP~\cite{hamamci2024foundation} & 69.4$\pm$2.2 & 64.4$\pm$1.6 & 68.5$\pm$1.4 & 30.3$\pm$1.5 & 63.9$\pm$1.6 & 67.1$\pm$1.8 \\
    fVLM~\cite{shui2025large} & 67.7$\pm$5.9 & 64.1$\pm$5.8 & 67.7$\pm$5.3 & 30.9$\pm$3.2 & 63.4$\pm$6.3 & 66.8$\pm$3.5 \\
    ViSD-Boost~\cite{cao2025boosting} & 63.9$\pm$9.8 & 61.8$\pm$7.5 & 65.9$\pm$6.9 & 29.3$\pm$5.2 & 61.1$\pm$8.3 & 64.2$\pm$7.6 \\
    Ours & \textbf{75.8$\pm$2.7} & \textbf{70.7$\pm$1.7} & \textbf{73.8$\pm$1.6} & \textbf{36.4$\pm$1.6} & \textbf{70.3$\pm$1.6} & \textbf{71.8$\pm$2.5} \\
    \midrule
    \multicolumn{7}{c}{\textbf{Rad-ChestCT}} \\
    \midrule
    fVLM~\cite{shui2025large} & 64.0$\pm$4.9 & 61.9$\pm$3.6 & 66.5$\pm$3.2 & 31.4$\pm$2.6 & 61.3$\pm$3.1 & 61.3$\pm$4.3 \\
    ViSD-Boost~\cite{cao2025boosting} & 58.6$\pm$7.5 & 56.6$\pm$5.7 & 61.7$\pm$5.1 & 28.1$\pm$3.8 & 56.6$\pm$6.2 & 57.8$\pm$5.3 \\
    Ours & \textbf{66.1$\pm$2.6} & \textbf{63.5$\pm$1.3} & \textbf{67.8$\pm$1.3} & \textbf{33.4$\pm$1.3} & \textbf{63.2$\pm$1.2} & \textbf{62.9$\pm$2.5} \\
    \bottomrule
  \end{tabular}
\label{tab:prompt_noncanonical}
\end{table*}

First, on the \textbf{Canonical} templates (Table~\ref{tab:prompt_canonical}), fVLM and ViSD-Boost perform well as these expressions are common in pre-training corpora, reaching 76.1--76.4 AUC on CT-RATE. However, CA-GCL achieves a higher mean AUC (78.2) with lower variance (std 0.7 vs.\ 2.7 and 1.8), indicating improved stability even when templates align with pre-trained distributions.

Second, on the \textbf{Non-canonical} templates (Table~\ref{tab:prompt_noncanonical}), fVLM and ViSD-Boost degrade notably, while CA-GCL remains stable with only a modest increase in standard deviation. On CT-RATE, fVLM drops to 67.7 (std 5.9) and ViSD-Boost to 63.9 (std 9.8), compared to CA-GCL's 75.8 (std 2.7). The gap widens on Rad-ChestCT, where ViSD-Boost falls to 58.6 (std 7.5) and fVLM to 64.0 (std 4.9), while CA-GCL maintains 66.1 (std 2.6). This suggests that non-conventional phrasing---such as \emph{Negative for''}, \emph{is not present''}, and \emph{``Ruled out''}---poses a greater challenge to baseline methods.

Third, the performance gap stems from differing sensitivities to syntactic patterns. Canonical templates follow a consistent structure that aligns with pre-trained distributions, whereas Non-canonical templates introduce lexical variations that require broader semantic understanding. By enforcing cross-anatomy separation at the global level, CA-GCL provides a more stable embedding space that is less sensitive to such variations. Overall, CA-GCL achieves the best mean AUC in both template groups on both datasets, demonstrating robustness to the linguistic diversity present in clinical practice.

\subsection{Analyses}

\subsubsection{Embedding Space Manifold Analysis}

\begin{figure}[t]
\centering
\includegraphics[width=0.95\textwidth]{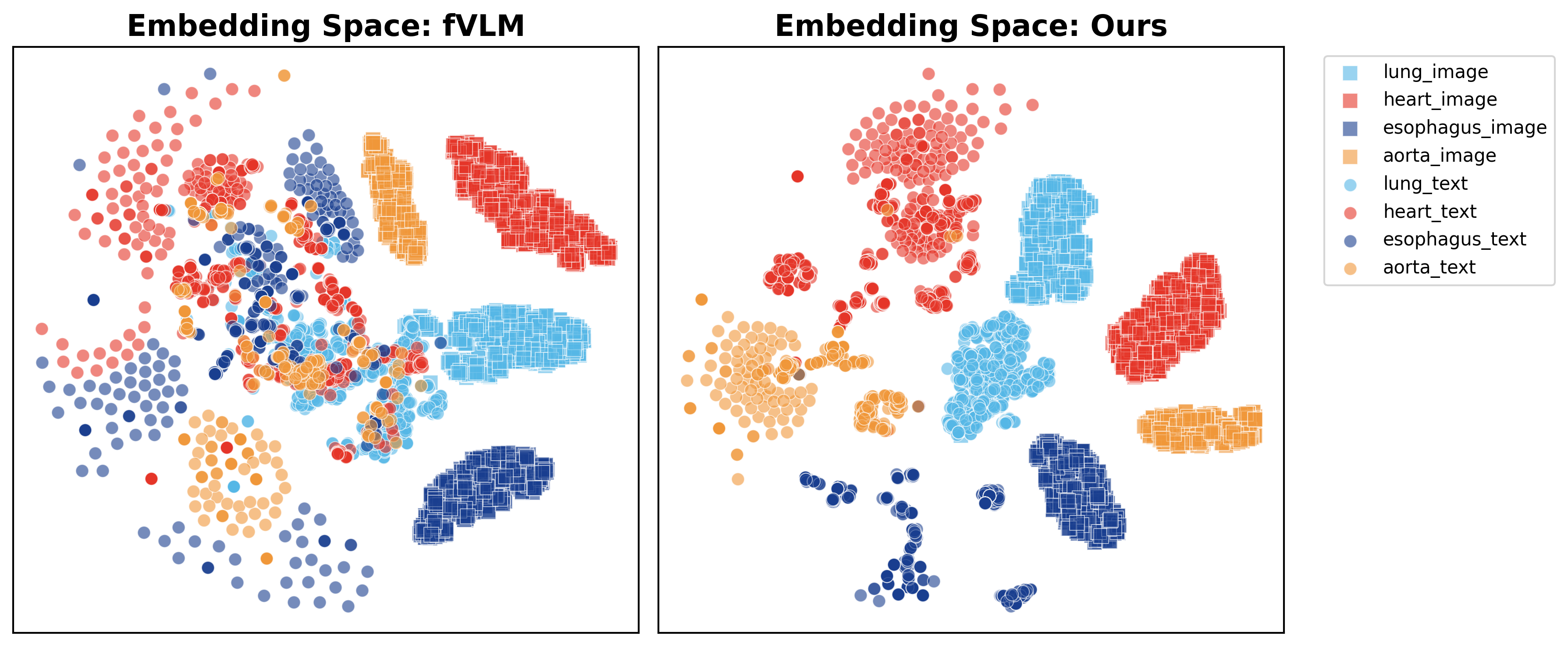}
\caption{t-SNE visualization of image and text embeddings for fVLM (left) and our method (right). While fVLM exhibits significant text embedding collapse, our method effectively separates anatomical clusters in the embedding space.}
\label{fig:4}
\end{figure}

We utilize t-SNE visualization to investigate the distribution of image and text embeddings within the embedding space. As shown in Fig.~\ref{fig:4} (left), fVLM exhibits severe representation collapse, where the circles representing different anatomical categories are heavily overlapped in various colors and remain indistinguishable. In contrast, as shown in Fig.~\ref{fig:4} (right), our method produces a well-structured manifold where anatomical clusters are clearly separated, validating the effectiveness of our proposed alignment strategy in learning discriminative representations.

\begin{table}[t]
  \centering
  \small
  \caption{Ablation study on the CT-RATE dataset. We evaluate the individual and cumulative efficacy of Anatomy-aware Local Contrastive Alignment (LCA), Clinical-Aware Text Augmentation (CTA), Report Jointly Parsing (RJP), and Cross-Anatomy Global Contrastive Alignment (GCA). Performance is measured by the mean and standard deviation across \textbf{10 zero-shot prompt templates} (5 Canonical and 5 Non-canonical, see Appendix~\ref{app:prompt_grouping}) to assess zero-shot robustness.}
  \begin{tabular}{cccc|cccccc}
    \toprule
    \multicolumn{4}{c|}{Module} & \multirow{2}{*}{AUC} & \multirow{2}{*}{ACC} & \multirow{2}{*}{F1} & \multirow{2}{*}{Prec} & \multirow{2}{*}{Spec} & \multirow{2}{*}{Sens} \\
    LCA & CTA & RJP & GCA & & & & & & \\
    \midrule
    \ding{51} & \ding{53} & \ding{53} & \ding{53} & 66.4$\pm$6.5 & 63.2$\pm$5.5 & 66.8$\pm$6.0 & 29.9$\pm$2.8 & 62.6$\pm$6.9 & 66.0$\pm$3.2 \\
    \ding{51} & \ding{53} & \ding{53} & \ding{51} & 73.3$\pm$2.6 & 68.5$\pm$1.8 & 71.8$\pm$1.7 & 34.9$\pm$1.6 & 68.5$\pm$1.5 & 70.1$\pm$2.4 \\
    \ding{51} & \ding{51} & \ding{53} & \ding{53} & 76.6$\pm$1.8 & 71.0$\pm$1.4 & 74.1$\pm$1.3 & 36.5$\pm$1.4 & 70.6$\pm$1.6 & 72.4$\pm$1.5 \\
    \ding{51} & \ding{51} & \ding{51} & \ding{53} & 76.7$\pm$1.1 & 71.2$\pm$0.7 & 74.3$\pm$0.6 & 36.4$\pm$0.7 & 71.2$\pm$0.8 & 71.8$\pm$1.2 \\
    \ding{51} & \ding{51} & \ding{51} & \ding{51} & \textbf{77.0$\pm$2.3} & \textbf{71.6$\pm$1.5} & \textbf{74.7$\pm$1.4} & \textbf{37.2$\pm$1.4} & \textbf{71.2$\pm$1.6} & \textbf{72.6$\pm$2.1} \\
    \bottomrule
  \end{tabular}
\label{tab:exp2}
\end{table}

\subsubsection{Ablation Study}

We report the mean and standard deviation of each metric computed over all 10 zero-shot prompt templates covering both Canonical and Non-canonical radiological patterns (Appendix~\ref{app:prompt_grouping}). Starting from the LCA-only baseline, incorporating either GCA or CTA yields substantial gains in AUC and F1-score, along with reduced variance, indicating improved robustness to prompt variations. Notably, the variant using only CTA exhibits the highest sensitivity, as the partial completeness strategy compels the model to identify abnormalities from sparse or incomplete textual descriptions, making it more sensitive to subtle pathological signs. Moreover, the performance gains from using GCA or CTA alone are smaller than those achieved when both are combined. We attribute this to a saturation effect in textual representation dispersion, where reducing severe embedding aggregation significantly improves prompt robustness, while further dispersion yields diminishing returns. Nevertheless, their combination still yields incremental improvements, suggesting the two methods are functionally complementary rather than redundant and validating the necessity of employing both for optimal performance.

\section{Conclusion}

In this paper, we identify and address the critical issue of representation collapse in FVLP. We introduce CA-GCL, which regularizes the geometric structure of the embedding space through a cross-anatomy global contrastive objective and clinical-aware text augmentation. Our theoretical motivation is validated by qualitative visualizations: t-SNE and cosine similarity distribution confirm that CA-GCL effectively separates anatomical clusters and produces a more balanced latent manifold. Extensive experimental results on the CT-RATE and Rad-ChestCT datasets demonstrate that CA-GCL achieves competitive zero-shot performance compared to existing FVLP paradigms, while exhibiting substantially stronger robustness to prompt variations. By resolving the distributional degeneracy inherent in previous fine-grained paradigms, CA-GCL takes a step toward more reliable AI assistants for 3D clinical diagnosis.

\bibliographystyle{splncs04}
\bibliography{ref}

\appendix

\section{Prompt Template Grouping Criteria}
\label{app:prompt_grouping}

To analyze prompt robustness, we partition the 10 zero-shot prompt templates into two groups of five (Table~\ref{tab:prompt_groups}). The grouping is determined by a joint criterion of \emph{empirical stability} and \emph{linguistic structure}: we cluster templates according to whether they conform to the canonical radiological negation syntax.

\textbf{Canonical templates} (Table~\ref{tab:prompt_groups}, upper) adhere to high-frequency radiological negation syntax, such as the ``No + abstract noun + of'' construction (\emph{e.g.}, \emph{``No findings of''}, \emph{``No diagnosis of''}, \emph{``No manifestation of''}), the nominalized antonym pair \emph{``Absence of / Presence of''}, and the direct negation \emph{``Not \{disease\}''}. These templates share a fixed syntactic frame that anchors the text encoder to familiar distributional modes in pre-trained language models.

\textbf{Non-canonical templates} (Table~\ref{tab:prompt_groups}, lower) deviate from this standard syntax, encompassing phrasal verbs (\emph{``Ruled out''}), adjective-predicate structures (\emph{``Negative for''}), colloquial multi-sense expressions (\emph{``No sign of''}), subject-predicate inversion (\emph{``is not present''}), and compound noun phrases (\emph{``No imaging findings of''}). These forms break the syntactic regularity of canonical negations and introduce semantic ambiguity that challenges models lacking explicit global regularization.

\begin{table}[h]
\centering
\scriptsize
\setlength{\tabcolsep}{4pt}
\caption{Grouping of the 10 zero-shot prompt templates into Canonical and Non-canonical sets.}
\begin{tabular}{ccll}
\toprule
\textbf{Group} & \textbf{ID} & \textbf{Negative Template} & \textbf{Positive Template} \\
\midrule
\multirow{5}{*}{Canonical} 
& 0 & Not \texttt{\{disease\}} & \texttt{\{disease\}} \\
& 1 & No findings of \texttt{\{disease\}} & Findings suggestive of \texttt{\{disease\}} \\
& 2 & Absence of \texttt{\{disease\}} & Presence of \texttt{\{disease\}} \\
& 3 & No diagnosis of \texttt{\{disease\}} & Diagnosis of \texttt{\{disease\}} \\
& 4 & No manifestation of \texttt{\{disease\}} & Manifestation of \texttt{\{disease\}} \\
\midrule
\multirow{5}{*}{Non-canonical} 
& 5 & Ruled out \texttt{\{disease\}} & Confirmed \texttt{\{disease\}} \\
& 6 & Negative for \texttt{\{disease\}} & Positive for \texttt{\{disease\}} \\
& 7 & No sign of \texttt{\{disease\}} & Sign of \texttt{\{disease\}} \\
& 8 & \texttt{\{disease\}} is not present & \texttt{\{disease\}} is present \\
& 9 & No imaging findings of \texttt{\{disease\}} & Imaging findings of \texttt{\{disease\}} \\
\bottomrule
\end{tabular}
\label{tab:prompt_groups}
\end{table}

\section{Per-Template Performance Breakdown}
\label{app:per_template}

To complement the grouped statistics in Table~\ref{tab:prompt_canonical} and Table~\ref{tab:prompt_noncanonical}, we report the full per-template zero-shot performance for all evaluated models. Table~\ref{app:tab:canonical_merged} details the results on the five \textbf{Canonical} templates (ID 0--4), and Table~\ref{app:tab:noncanonical_merged} details the results on the five \textbf{Non-canonical} templates (ID 5--9). 

\begin{table*}[t]
\scriptsize
\centering
\setlength{\tabcolsep}{1.5pt}
\caption{Per-template zero-shot performance on \textbf{Canonical} prompt templates (ID 0, 1, 2, 3, 4). The best result in each column per dataset is \textbf{bold}. ``---'' indicates the model was not evaluated on the respective dataset.}
\begin{tabular}{cc|cccccc|cccccc}
\toprule
\multirow{2}{*}{\textbf{ID}} & \multirow{2}{*}{\textbf{Method}} & \multicolumn{6}{c|}{\textbf{CT-RATE}} & \multicolumn{6}{c}{\textbf{Rad-ChestCT}} \\
\cmidrule(lr){3-8} \cmidrule(lr){9-14}
& & AUC & ACC & F1 & Prec & Spec & Sens & AUC & ACC & F1 & Prec & Spec & Sens \\
\midrule
\multirow{4}{*}{0} 
& CT-CLIP & 72.7 & 66.5 & 70.5 & 32.3 & 65.9 & 70.0 & --- & --- & --- & --- & --- & --- \\
& fVLM & 76.4 & 70.9 & 73.9 & 36.2 & 70.4 & 72.7 & 65.7 & 62.0 & 66.5 & 32.2 & 61.6 & 63.1 \\
& ViSD-Boost & \textbf{79.5} & \textbf{73.5} & \textbf{76.3} & \textbf{38.8} & \textbf{73.1} & 74.9 & \textbf{69.8} & 65.3 & 69.5 & \textbf{34.5} & 64.9 & \textbf{66.9} \\
& Ours & 79.3 & 72.7 & 75.7 & 38.4 & 71.9 & \textbf{75.5} & 67.7 & \textbf{65.5} & \textbf{69.7} & 34.3 & \textbf{65.6} & 62.7 \\
\cmidrule(lr){1-14}
\multirow{4}{*}{1} 
& CT-CLIP & 71.6 & 66.0 & 70.0 & 31.7 & 65.2 & 69.6 & --- & --- & --- & --- & --- & --- \\
& fVLM & 77.7 & 71.2 & 74.4 & 36.7 & 70.7 & 73.9 & 67.5 & 63.6 & 68.1 & 33.3 & 63.8 & 63.9 \\
& ViSD-Boost & \textbf{78.1} & 71.9 & 74.8 & \textbf{37.5} & 70.6 & \textbf{75.0} & \textbf{69.7} & \textbf{64.3} & \textbf{68.6} & \textbf{34.1} & \textbf{64.1} & \textbf{66.5} \\
& Ours & 77.5 & \textbf{72.2} & \textbf{75.3} & 37.4 & \textbf{71.8} & 72.6 & 65.1 & 62.8 & 67.2 & 32.8 & 62.6 & 61.6 \\
\cmidrule(lr){1-14}
\multirow{4}{*}{2} 
& CT-CLIP & 72.9 & 66.8 & 70.7 & 32.5 & 65.8 & 71.0 & --- & --- & --- & --- & --- & --- \\
& fVLM & 73.1 & 67.6 & 70.9 & 33.9 & 67.4 & 70.8 & 65.4 & 61.0 & 65.7 & 31.3 & 60.0 & 64.5 \\
& ViSD-Boost & 73.6 & 68.6 & 72.2 & 33.7 & 68.5 & 69.1 & 62.5 & 59.6 & 64.4 & 29.1 & 60.5 & 59.1 \\
& Ours & \textbf{77.8} & \textbf{71.6} & \textbf{74.8} & \textbf{37.3} & \textbf{71.2} & \textbf{73.5} & \textbf{68.7} & \textbf{63.6} & \textbf{67.9} & \textbf{34.4} & \textbf{63.1} & \textbf{66.5} \\
\cmidrule(lr){1-14}
\multirow{4}{*}{3} 
& CT-CLIP & 72.3 & 66.6 & 70.6 & 32.2 & 65.9 & 69.6 & --- & --- & --- & --- & --- & --- \\
& fVLM & \textbf{78.3} & 71.9 & 75.0 & 37.1 & 71.2 & 74.2 & 67.4 & \textbf{64.7} & \textbf{69.0} & \textbf{34.0} & \textbf{65.1} & 62.9 \\
& ViSD-Boost & 77.2 & 70.4 & 73.8 & 35.9 & 69.2 & \textbf{74.4} & \textbf{68.5} & 63.5 & 67.9 & 33.3 & 63.2 & \textbf{65.8} \\
& Ours & 77.9 & \textbf{73.0} & \textbf{75.9} & \textbf{38.4} & \textbf{73.2} & 72.5 & 66.5 & 63.8 & 68.2 & 33.5 & 63.3 & 63.5 \\
\cmidrule(lr){1-14}
\multirow{4}{*}{4} 
& CT-CLIP & 70.2 & 64.5 & 68.7 & 30.9 & 63.6 & 69.1 & --- & --- & --- & --- & --- & --- \\
& fVLM & 76.7 & 71.0 & 74.0 & 36.0 & 70.8 & 72.4 & \textbf{69.2} & \textbf{64.6} & \textbf{69.0} & 34.1 & 63.9 & \textbf{66.6} \\
& ViSD-Boost & 72.3 & 66.7 & 70.5 & 31.8 & 66.2 & 69.4 & 63.2 & 58.4 & 63.6 & 29.1 & 57.9 & 62.6 \\
& Ours & \textbf{78.7} & \textbf{72.9} & \textbf{75.8} & \textbf{38.3} & \textbf{72.9} & \textbf{73.3} & 67.8 & 64.4 & 68.7 & \textbf{34.3} & \textbf{64.3} & 64.3 \\
\bottomrule
\end{tabular}
\label{app:tab:canonical_merged}
\end{table*}

\begin{table*}[t]
\scriptsize
\centering
\setlength{\tabcolsep}{1.5pt}
\caption{Per-template zero-shot performance on \textbf{Non-canonical} prompt templates (ID 5, 6, 7, 8, 9). The best result in each column per dataset is \textbf{bold}. ``---'' indicates the model was not evaluated on the respective dataset.}
\begin{tabular}{cc|cccccc|cccccc}
\toprule
\multirow{2}{*}{\textbf{ID}} & \multirow{2}{*}{\textbf{Method}} & \multicolumn{6}{c|}{\textbf{CT-RATE}} & \multicolumn{6}{c}{\textbf{Rad-ChestCT}} \\
\cmidrule(lr){3-8} \cmidrule(lr){9-14}
& & AUC & ACC & F1 & Prec & Spec & Sens & AUC & ACC & F1 & Prec & Spec & Sens \\
\midrule
\multirow{4}{*}{5} 
& CT-CLIP & 66.0 & 62.1 & 66.5 & 27.8 & 61.9 & 63.9 & --- & --- & --- & --- & --- & --- \\
& fVLM & 69.8 & 69.3 & 72.4 & 33.0 & 70.1 & 64.6 & \textbf{68.9} & \textbf{65.7} & \textbf{69.9} & \textbf{34.5} & \textbf{65.1} & \textbf{65.3} \\
& ViSD-Boost & 68.3 & \textbf{69.8} & \textbf{73.0} & 33.2 & \textbf{71.9} & 61.0 & 66.9 & 62.7 & 67.3 & 32.2 & 62.7 & 63.9 \\
& Ours & \textbf{73.8} & 69.3 & 72.6 & \textbf{35.3} & 69.1 & \textbf{69.8} & 64.6 & 63.0 & 66.9 & 32.7 & 62.7 & 61.7 \\
\cmidrule(lr){1-14}
\multirow{4}{*}{6} 
& CT-CLIP & 68.8 & 63.7 & 67.9 & 29.9 & 63.1 & 67.3 & --- & --- & --- & --- & --- & --- \\
& fVLM & 57.3 & 53.5 & 58.2 & 25.0 & 52.3 & 61.5 & 54.7 & 55.0 & 60.3 & 26.7 & 55.6 & 52.9 \\
& ViSD-Boost & 64.7 & 61.8 & 66.1 & 29.4 & 61.1 & 65.7 & 51.8 & 52.0 & 57.8 & 24.5 & 52.8 & 50.5 \\
& Ours & \textbf{76.3} & \textbf{71.0} & \textbf{74.2} & \textbf{36.7} & \textbf{70.3} & \textbf{73.1} & \textbf{68.1} & \textbf{64.8} & \textbf{69.0} & \textbf{34.2} & \textbf{63.6} & \textbf{65.2} \\
\cmidrule(lr){1-14}
\multirow{4}{*}{7} 
& CT-CLIP & 69.0 & 64.4 & 68.5 & 30.4 & 63.7 & 66.8 & --- & --- & --- & --- & --- & --- \\
& fVLM & \textbf{72.5} & 67.1 & 70.6 & 32.9 & 66.3 & \textbf{70.6} & \textbf{65.5} & \textbf{62.7} & \textbf{67.2} & \textbf{31.7} & \textbf{61.8} & \textbf{62.9} \\
& ViSD-Boost & 65.7 & 61.0 & 64.8 & 29.9 & 59.4 & 68.9 & 55.7 & 54.2 & 59.3 & 26.6 & 53.6 & 57.5 \\
& Ours & 71.7 & \textbf{68.2} & \textbf{71.5} & \textbf{34.0} & \textbf{67.9} & 68.0 & 62.0 & 61.2 & 65.7 & 31.3 & 61.5 & 58.6 \\
\cmidrule(lr){1-14}
\multirow{4}{*}{8} 
& CT-CLIP & 72.8 & 67.0 & 70.9 & 32.3 & 66.8 & 69.1 & --- & --- & --- & --- & --- & --- \\
& fVLM & 65.5 & 62.3 & 65.9 & 30.2 & 60.8 & 66.8 & 64.5 & 62.4 & 67.0 & 32.1 & 61.2 & 62.5 \\
& ViSD-Boost & 45.8 & 48.4 & 53.7 & 19.6 & 46.9 & 51.5 & 50.5 & 49.9 & 55.7 & 24.3 & 48.8 & 53.5 \\
& Ours & \textbf{78.6} & \textbf{72.3} & \textbf{75.3} & \textbf{37.9} & \textbf{72.0} & \textbf{73.8} & \textbf{69.2} & \textbf{64.7} & \textbf{69.0} & \textbf{35.1} & \textbf{65.2} & \textbf{64.8} \\
\cmidrule(lr){1-14}
\multirow{4}{*}{9} 
& CT-CLIP & 70.3 & 64.7 & 68.9 & 31.1 & 63.9 & 68.4 & --- & --- & --- & --- & --- & --- \\
& fVLM & 73.3 & 68.1 & 71.6 & 33.6 & 67.3 & 70.5 & 66.2 & 63.6 & 67.9 & 32.2 & 62.8 & 63.0 \\
& ViSD-Boost & 75.1 & 68.1 & 71.8 & 34.3 & 66.3 & 74.0 & \textbf{68.1} & \textbf{64.1} & \textbf{68.3} & 33.1 & \textbf{65.1} & 63.4 \\
& Ours & \textbf{78.5} & \textbf{72.5} & \textbf{75.5} & \textbf{38.1} & \textbf{72.1} & \textbf{74.3} & 66.8 & 63.9 & \textbf{68.3} & \textbf{33.8} & 62.8 & \textbf{64.2} \\
\bottomrule
\end{tabular}
\label{app:tab:noncanonical_merged}
\end{table*}

\end{document}